\documentclass[10pt,twocolumn,letterpaper]{article}

\usepackage[pagenumbers]{cvpr}

\usepackage{float}

\definecolor{cvprblue}{rgb}{0.21,0.49,0.74}
\usepackage[pagebackref,breaklinks,colorlinks,allcolors=cvprblue]{hyperref}

\title{A Mechanistic Analysis of Adversarial Fine-tuning of Vision Transformers}

\author{Hannah Gao\thanks{Equal contribution.}\\
MIT\\
Cambridge, MA\\
{\tt\small hanngao@mit.edu}
\and
Isha Agarwal\footnotemark[1]\\
MIT\\
Cambridge, MA\\
{\tt\small agarwali@mit.edu}
\and
Dylan Hadfield-Menell\\
MIT\\
Cambridge, MA\\
{\tt\small dylanhm@mit.edu}
\and
Rachel Ma\\
MIT\\
Cambridge, MA\\
{\tt\small rachelm8@mit.edu}
}

\begin{document}
\maketitle
\begin{abstract}
The widespread use of image classification models in high-risk, real-world situations necessitates making these models robust to slight disturbances or perturbations, such as blurring or sharpening, in the input images. While vision transformers (ViTs) play an integral role in many modern-day multi-modal models like Vision-Language-Models (VLMs) and Vision-Language-Action (VLA) models, they have received a lack of attention in the setting of robustness. In this work, we analyze the effects of adversarial fine-tuning, a popular method for improving model robustness to image perturbations, on a ViT's performance on perturbed and regular images through a mechanistic lens. We adversarially train a ViT on low-frequency and high-frequency image corruptions, and attempt to explain changes in downstream model performance through an examination of the model's attention mechanisms, internal representations, and knowledge evolution. Overall, our results suggest that, while fine-tuning on inputs with common corruptions improves model performance and certainty on new instances of corrupted data, these improvements do not transfer to other classes of corruptions not seen in the training. Additionally, despite observing changes in visual attention and knowledge evolution across layers, we found that adversarial training did not lead to fundamental changes in the sparse representations learned by ViTs.
\end{abstract}    
\section{Introduction}
\label{sec:intro}

Image classification models have wide-ranging applications, from medical image analysis to self-driving cars \cite{xin2025med3dvlmefficientvisionlanguagemodel, ahammed2024computervisionapproachautonomous}. It is important to develop classification models invariant to image perturbations, including minor modifications such as blurs, that commonly occur in the real world. For example, security cameras should continue to work to classify threats, even when there is fog or other weather elements that may interfere with the lens's ability to get a clear photo. Additionally, given the growing popularity of vision-language models (VLMs) and vision-language-action (VLA) models, it is more important than ever to study the robustness of vision models, like Vision Transformers (ViTs), which are directly being augmented into such models for perception \cite{kim2024openvlaopensourcevisionlanguageactionmodel}.

One way of making models more robust is training on more adversarial examples.  We look at the effect of adversarial training on simple and common image corruptions. It is important to understand exactly how adversarial training is influencing the model, mechanistically, as fine-tuning has been shown to have inadvertent side-effects on learning and performance, such as catastrophic forgetting \cite{zhai2023investigatingcatastrophicforgettingmultimodal}.

Our project investigates how robust latent representations of images are to image perturbations in the form of common image corruptions. We examine how the internal activations of a Vision Transformer (ViT) image classification model change when given an image versus its corrupted counterpart. We also explore whether fine-tuning the ViT on perturbed images improves robustness of the model's learned features.

We explore the effects of adversarially training a ViT on corrupted images through two main questions:
1) What is the impact of adversarial training on downstream ViT performance? and 2) How can we explain the changes in performance mechanistically?

To answer these questions, we compare how the way models process perturbed and original images differ before and after adversarial fine-tuning on different image corruptions. To this end, we examine how different the model attentions are for blurred versus original images in each model. Then, we train a Sparse Autoencoder (SAE), which provides us with interpretable feature representations, on each model to understand how internal representations of blurred versus original images are affected by adversarial training.

Our main contributions can be summarized as follows: 

\begin{enumerate}
    \item We fine-tune ViT models on two different families of common corruptions and compare downstream performance and find that adversarially fine-tuning on low frequency perturbations does not translate to improved performance on high-frequency perturbations.
    \item We apply interpretability techniques to understand how the adversarial fine-tuning changes the model both in its attention mechanism and the latent representations. 
    \item We observe that adversarially fine-tuned models exhibit greater confidence in correct answers and the emergence of correct classifications earlier in during input processing.      
\end{enumerate}

\section{Related Works}

\textbf{Robustness of Vision Models.} 
While in recent years vision models have demonstrated impressive capabilities across tasks ranging from object-identification to spatial analysis, they still suffer performance degradations when given images with some form of perturbation. 

Perturbations consist of adversarial attacks, which are perturbations specially engineered to fool the model, or common corruptions \cite{hendrycks2019benchmarkingneuralnetworkrobustness} such as blurs or noise, which resemble more natural perturbations that may occur in the real world.

While most works have focused on adversarial attacks on vision models \cite{rowe2022closerlookrobustnesslinfinity, goodfellow2015explainingharnessingadversarialexamples, madry2019deeplearningmodelsresistant}, several studies have also examined the the effects of common image corruptions and have found vision models to be vulnerable to various degrees to several common corruption types, particularly  to high-frequency perturbations \cite{usama2025analysingrobustnessvisionlanguagemodelscommon, wang2024surveyrobustnesscomputervision}. However, there has been relatively little work to analyze the robustness of ViTs to image perturbations \cite{mahmood2021robustnessvisiontransformersadversarial} and particularly to these common corruptions. Furthermore, to our knowledge, such works only focus simply on downstream performance impacts of common corruptions and propose mitigations like adversarial training \cite{usama2025analysingrobustnessvisionlanguagemodelscommon, ghosh2025cleadcontrastivelearningenhanced} without going any further to mechanistically understand the underlying reason for observed performance changes. 

\textbf{Mechanistic Interpretability on Vision Models.} While most prior mechanistic interpretability works have focused on language models \cite{wang2022interpretabilitywildcircuitindirect, cunningham2023sparseautoencodershighlyinterpretable, meng2023locatingeditingfactualassociations}, more recent literature has started to adapt techniques for language models such as examining activations, logit attribution and training SAEs for vision models \cite{jiang2025visiontransformersdontneed, joseph2024visionmechinterp, stevens2025interpretabletestablevisionfeatures}. There have also been works that have come up with or applied vision-specific interpretability techniques \cite{Selvaraju_2019, Toker_2024}.

However, the internal workings of vision models remain understudied, especially in the context of fine-tuning and robustness. Given that robustness to perturbations is a crucial property for any vision model deployed in the real world where images may naturally be perturbed, we are interested in studying and quantifying this phenomenon by studying the model's internal processing of perturbed images.

\section{Methods}

\begin{figure}[t]
    \centering
    \includegraphics[width=1.0\linewidth]{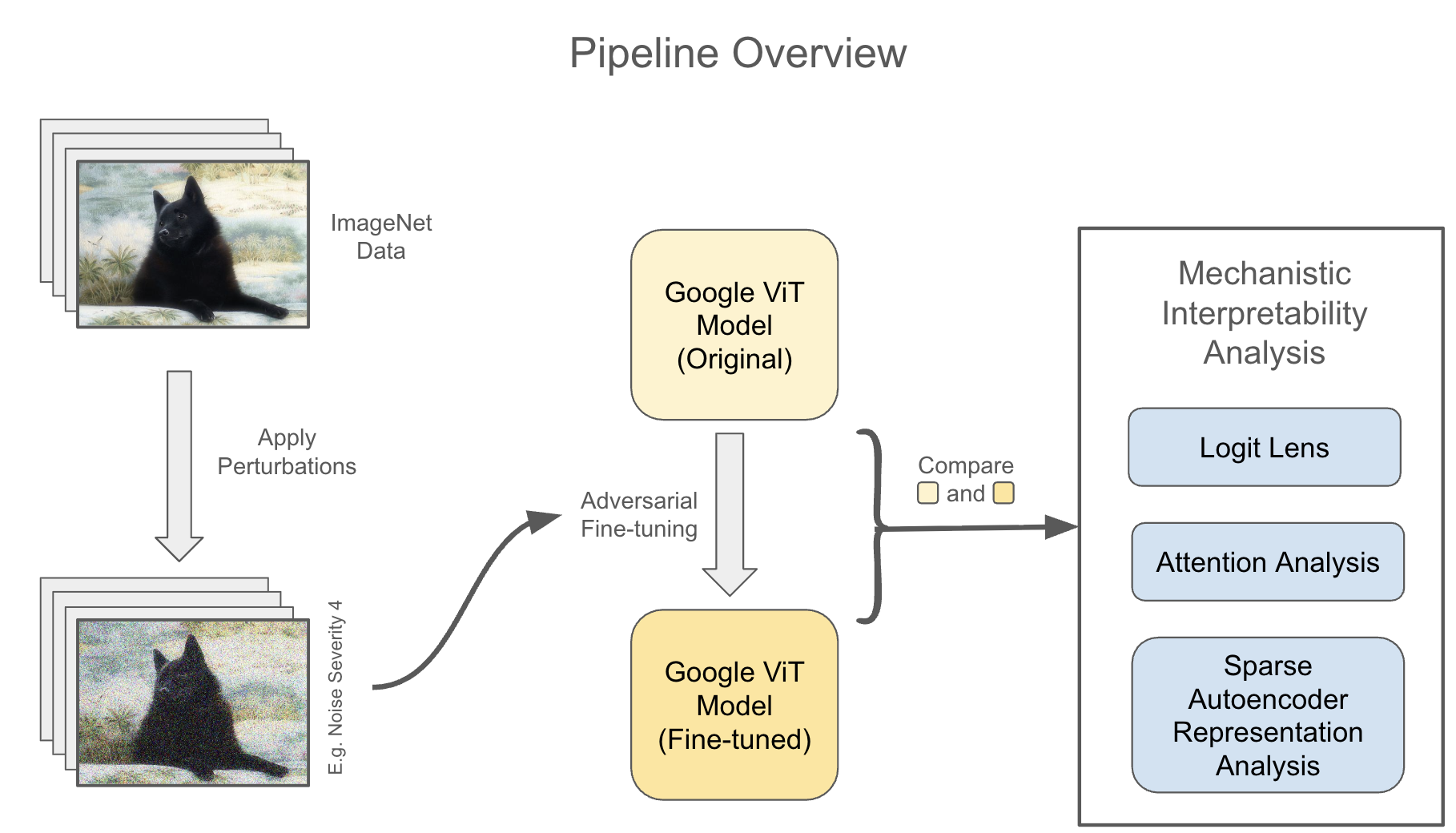}
    \caption{Overview of methods. ViT is adversarially trained on perturbed ImageNet images, and the fine-tuned and original base ViT are analyzed and compared using various mechanistic interpretability techniques. Images sourced (and modified) from ImageNet ILSVRC2012 dataset.}
    \label{fig:method_overview}
\end{figure}
We apply common corruptions to an image dataset and use these perturbed images to adversarially fine-tune a ViT. In addition to comparing the base and fine-tuned models on downstream performance on new perturbed images, we additionally compare the two models using a suite of mechanistic interpretability techniques (see \cref{fig:method_overview})

\subsection{Adversarial Training}

We perform adversarial training by fine-tuning a ViT on image corruptions on the ImageNet ILSVRC2012 dataset \cite{5206848}. We broadly explore two main categories of image corruptions: low-frequency and high-frequency corruptions. For our low-frequency corruption, we apply the common Gaussian blur, and for our high-frequency corruption, we apply a Gaussian noise filter. We apply these filters each at severity levels $1$, $2$, and $4$, where for the Gaussian blur the level is used as the standard deviation of the blur kernel, while for Gaussian noise we follow the implementation of the severity levels used in ImageNetC \cite{hendrycks2019benchmarkingneuralnetworkrobustness}. In both cases, a higher severity level indicates a higher degree of image corruption.

\subsection{Class Prediction Progression Across Layers}

To understand how the model's class prediction evolves across layers, we apply the logit lens technique \cite{nostalgebraist2020logitlens}. For LLMs, this techniques applies the unembedding matrix matrix $W_U \in \mathbb{R}^{d_{model} \times |V|}$ to project an intermediate hidden state $h^{(l)} \in \mathbb{R}^{d_{model}}$, producing $\text{logits}^{(l)}$ over the vocabulary $V$. Taking the token with the maximum logit in $\text{logits}^{(l)}$ then allows us to determine what the model ``predicts'' the next token to be at that layer. 

We modify the technique for a ViT classifier by examining the classifier head output when applied to $c^{(l)}$, the CLS token representation at an intermediate layer $l$, to produce $\text{logits}^{(l)}$. We then apply a softmax to get the model's probability distribution over the possible classes for that layer.

For inputs on which the model considers the correct classification at some point in its processing, we quantify the layer at which the correct answer first emerges as a prediction. Similar to logit lens for LLMs, we take the model's class prediction at layer $l$ to be the maximum logit after applying the classifier head to $c^{(l)}$.

\subsection{Attention Analysis}

We consider how the model's underlying attention structure changes on perturbed inputs. Let $A^{(l)}_{h, i, j}$ denote the attention output of head $h$ in layer $l$ for query token $i$ attending to token $j$. 

We measure how close attentions for the original versus perturbed input are at each layer by examining squared difference and cosine similarity. To quantify how perturbation impacts attention distribution over the token, we also compute the entropy $H^{(l)}_{h, i}$:
\begin{equation}
    H^{(l)}_{h, i} = - \sum_{j = 1}^T A^{(l)}_{h, i, j} \log(A^{(l)}_{h, i, j}).
\end{equation}
We then take an average over all heads and query tokens to produce average entropy $\overline{H}^{(l)}$ for each layer. We compare $\overline{H}^{(l)}$ for an original image $x$ and a perturbed image $\tilde{x}$ to see how perturbation affects attention spread over the image patches.

\subsection{Representational Analysis}
To understand how learned representations in the ViT are affected by adversarial training, we explore the internal activations of the ViT.  To disentangle the internal representations of the model into a dictionary of sparse concepts, we train Sparse Autoencoders (SAEs) on the outputs of the penultimate ViT transformer layer, following guidance from previous works such as \cite{gao2025scaling} which suggest placing SAEs close to the end of the model.

An SAE encodes representations into a higher-dimensional representation space and then decodes back to the original dimension, with the objective of reconstructing representations as closely as possible while simultaneously enforcing sparsity in the training loss to ensure that the features in the SAE are sufficiently disentangled, i.e. that very few SAE neurons activate strongly on a given input. \\

We train both a Vanilla ReLU SAE \cite{cunningham2023sparseautoencodershighlyinterpretable}, on the loss:
\begin{equation}
    ||x- \text{Dec}(\text{Enc}(x))||^2_2 + \lambda||\text{Enc}(x)||_1
\end{equation}
and a BatchTopK SAE \cite{bussmann2024batchtopksparseautoencoders}, using the loss:
\begin{equation}
    ||x- \text{Dec}(z_{topk})||^2_2 + \lambda|| z_{topk}||_1 + \alpha L_{aux}
\end{equation}
where $z_{topk}=\text{BatchTopK}(\text{Enc}(x))$, and Enc and Dec are the typical encoding and decoding functions consisting of an encoding/decoding matrix respectively and a bias term. We use the BatchTopK function provided by \cite{bussmann2024batchtopksparseautoencoders}.

\section{Experiments}

All experiments are performed on Google's ViT-B/16 model \cite{dosovitskiy2021imageworth16x16words, wu2020visual, 5206848} which is a vision classifier pre-trained on ImageNet21 (containing 4 million images and 21k target classes) and fine-tuned on ImageNet ILSVRC2012 (containing 1 million images and 1k target classes) \cite{5206848, ILSVRC15}.

We apply each of our 6 perturbations (3 levels of Gaussian blur and 3 levels of Gaussian noise) on 10k images from held-out ImageNet validation data, and fine-tune the ViT separately on each of these datasets. We fine-tune for up to 10 epochs, using early-stopping, learning rate 5e-5, AdamW optimizer, and a linear learning-rate scheduler. 

\subsection{Adversarial Finetuning Results}

We evaluate all the adversarially trained models on 1k new blurred and noised image samples, and compare their performance with the original base ViT. 

We notice that the ViTs fine-tuned on Gaussian blurs resulted in increased downstream accuracy (i.e., top-1-accuracy), top-5 accuracy, and top-10 accuracy on classifying new blurred images, and the performance gain was monotonically increasing in the level of blur used in the adversarial training data (i.e., blur-level-4-tuned ViTs performed the best). (See \cref{tab:model_performance}) We noticed the same trend when evaluating the various levels of Gaussian-noise-tuned models on new noised image samples.

\begin{figure}[t]
    \centering
    \includegraphics[width=1.0\linewidth]{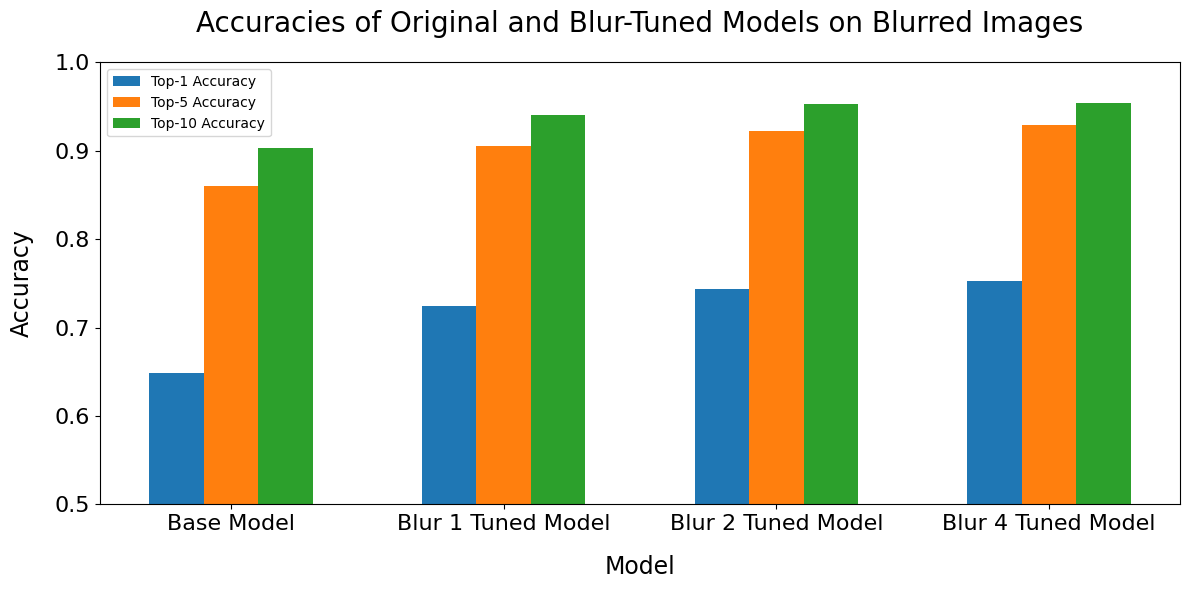}
    \caption{Top-1-accuracy, top-5-accuracy, and top-10 accuracy of the blur-tuned models on blur-4 test images.}
    \label{fig:blur_accuracies}
\end{figure}

\begin{figure}[t]
    \centering
    \includegraphics[width=1.0\linewidth]{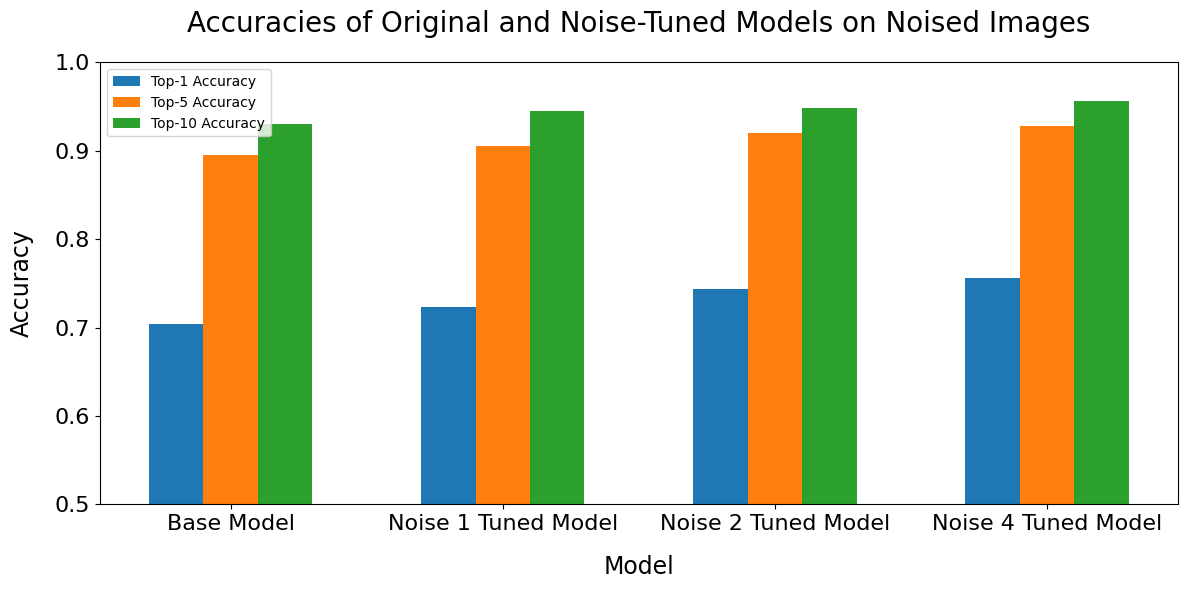}
    \caption{Top-1-accuracy, top-5-accuracy, and top-10 accuracy of the noise-tuned models on noise-4 test images.}
    \label{fig:noise_accuracies}
\end{figure}

Interestingly, we notice that across the blur-tuned models (i.e., models fine-tuned on Gaussian blur datasets), the metric that saw the greatest improvement on novel blurred data was top-1-accuracy, followed by top-5-accuracy, and finally top-10-accuracy; the same pattern was observed for noise-tuned models. (See \cref{fig:blur_accuracies} and \cref{fig:noise_accuracies}).

\begin{table}[t]
\centering
\begin{tabular}{|l|p{1cm}|p{1cm}|p{1cm}|p{0.7cm}|}\hline

Model & Acc.& Top 5 Acc.& Top 10 Acc.& ECE \\\hline

Base & 0.649 & 0.860 & 0.903 & 0.054 \\\hline
Blur 1 tuned & 0.725 & 0.905 & 0.941 & 0.046 \\\hline
Blur 2 tuned & 0.744 & 0.922 & 0.953 & 0.054 \\\hline

 Blur 4 tuned & 0.753& 0.929& 0.954&0.055\\ \hline
\end{tabular}
\caption{Comparison of accuracies and expected calibration error (ECE) for models fine-tuned on images with Gaussian blur severity levels 1, 2, and 4.}
\label{tab:model_performance}
\end{table}

These results suggest that adversarial training on higher severity of common corruptions like Gaussian blurs and Gaussian noise lead to increased model certainty in the correct classification on future similar corruptions: while all models seem to have similar rates of predicting the correct answer within their top-10 (and even top-5) answers, the models fine-tuned on higher levels of blur and noise filters had notably higher top-1 accuracies.

We observed minimal to no changes in the top-1, top-5, and top-10 accuracies for blur-tuned models evaluated on the noise-4 test data and for noise-tuned models evaluated on the blur-4 test data. This indicates that robustness gained from adversarial training on one family of corruptions is not transferrable, though not harmful, to robustness in other families of corruptions.

We also observe minimal to no changes in the top-1, top-5, and top-10 accuracies of the adversarially-trained models on uncorrupted ImageNet images, suggesting that moderate amounts of fine-tuning on corruptions may confer neither general performance benefits nor harm to ViTs.

\subsection{Class Prediction Progression Across Layers}
\begin{figure}[t]
    \centering
    \includegraphics[width=0.9\linewidth]{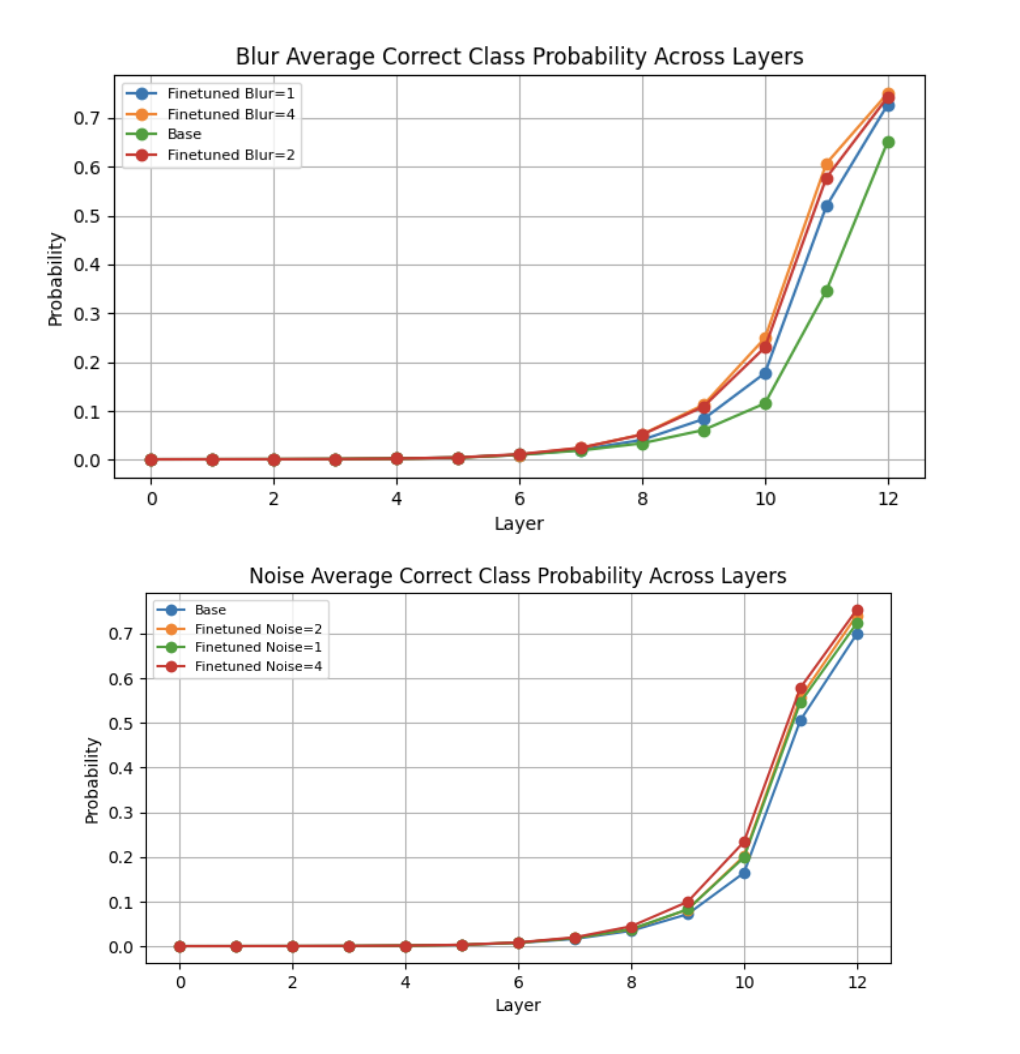}
    \caption{Probability of predicting the correct class for base and finetuned models on dataset with Gaussian blur (top) and Gaussian noise (bottom) of severity $4$ applied.}
    \label{fig:logit-attrib}
\end{figure}

We look at the average probability of predicting the correct class throughout the layers of each fine-tuned model. The results for both the Gaussian blur and noise are included in \cref{fig:logit-attrib}.

The results suggest that while all models start out with about the same probability of predicting the correct class, models fine-tuned on the highest filter severity (which also exhibited the best performance on these datasets) tend to overtake other models by having a higher probability of predicting the correct class in a mid-to-late layer. This suggests the mechanism for improved accuracy may be localized to those layers, or that these layers may be responsible for some difference in processing corrupted images, though further investigation is required to determine whether such a mechanism exists and may be controlled. 

\begin{figure}[t]
    \centering
    \includegraphics[width=0.9\linewidth]{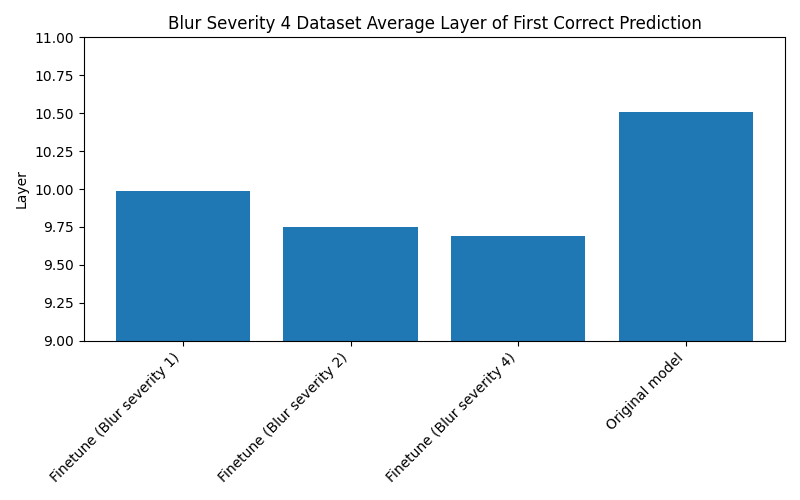}
    \caption{The average first layer the model predicts the correct class (restricted to only consider inputs on which the models predict the correct class at some layer) on the blur severity four dataset.}
    \label{fig:first_pred_blur}
\end{figure}

\begin{figure}[t]
    \centering
    \includegraphics[width=0.9\linewidth]{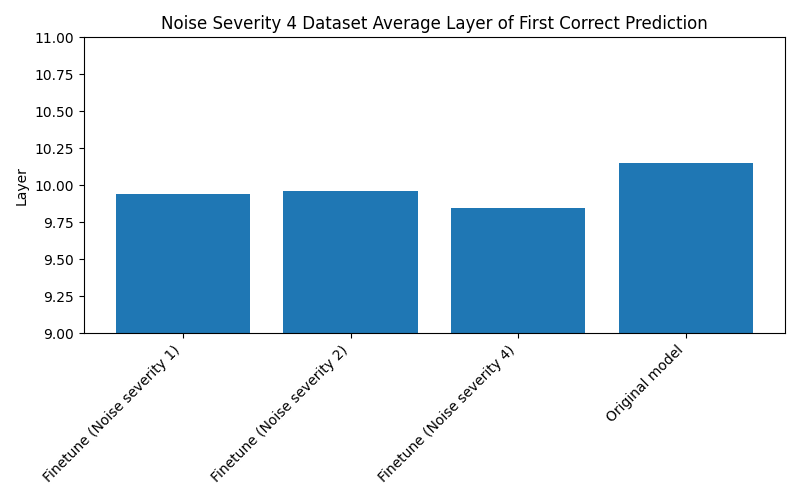}
    \caption{The average first layer the model predicts the correct class (restricted to only consider inputs on which the models predict the correct class at some layer) on the noise severity four dataset.}
    \label{fig:first_pred_noise}
\end{figure}

When restricting to inputs for which model  predicts the correct class at some layer, we find that the first layer in which the fine-tuned models learns (i.e. has the highest logit for) the correct class is lower than the original model on the adversarial datasets, as shown in \cref{fig:first_pred_blur} and \cref{fig:first_pred_noise}. The effect is more pronounced for Gaussian blur than for Gaussian noise. Overall, this suggests that after fine-tuning, ViTs are able to tell what the correct image classification is earlier on in processing for adversarial perturbations.  

\subsection{Attention Analysis}

\begin{figure}[t]
    \centering
    \includegraphics[width=0.9\linewidth]{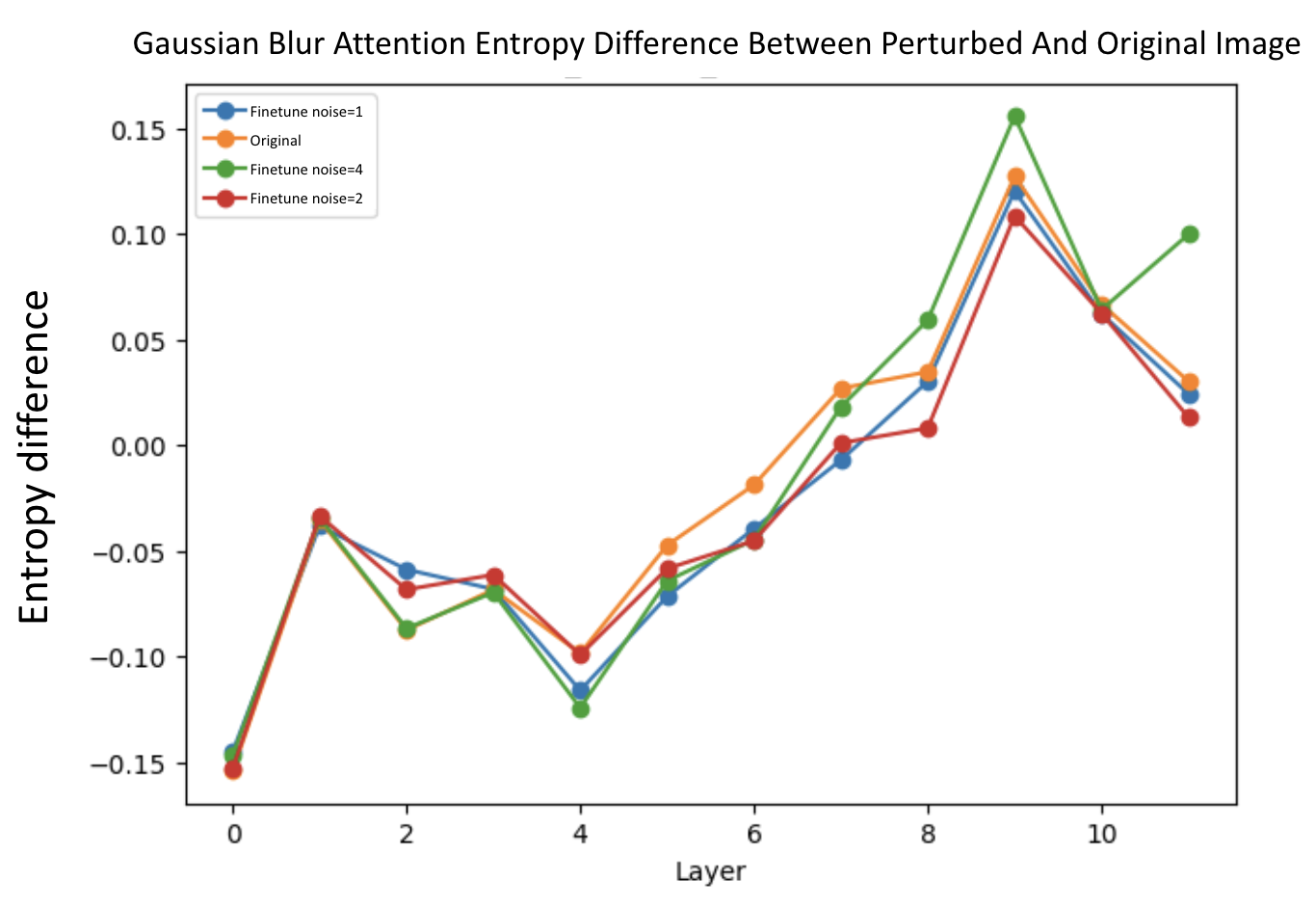}
    \caption{The average  difference in attention entropy between an original image and blurred image across the layers of the model.}
    \label{fig:attn_entropy_blur}
\end{figure}

\begin{figure}[t]
    \centering
    \includegraphics[width=0.9\linewidth]{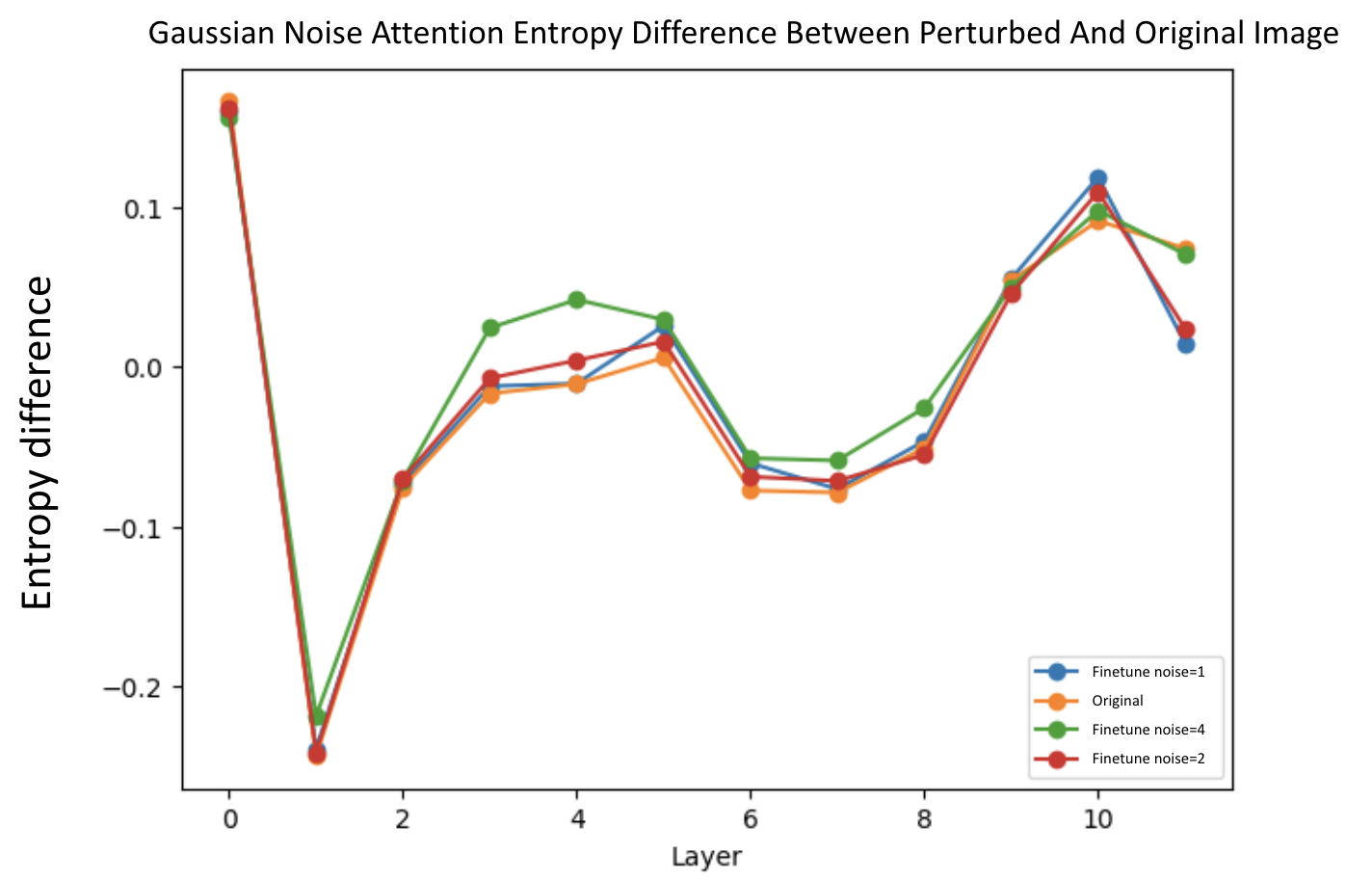}
    \caption{The average  difference in attention entropy between an original image and noised image across the layers of the model.}
    \label{fig:attn_entropy_noise}
\end{figure}

We look at the results of the difference in attention entropy between the original and blurred images for each model in \cref{fig:attn_entropy_blur} and \cref{fig:attn_entropy_noise}. For both perturbations, the difference starts out negative, suggesting that the perturbed image attentions are more centralized as opposed to spread out over the whole image than the original image attentions. However, as the model processes the image, the difference becomes positive. 

For the blur dataset in \cref{fig:attn_entropy_blur}, we observe that the difference is highest for the original model, suggesting that its blurred image attention entropy is lower and possibly more skewed towards certain image patches as opposed to being more evenly spread out. This suggests that robustness fine-tuning for Gaussian blurs may enable models to focus on more specific patches of the image as opposed to the overall image more broadly. 

On the other hand, for Gaussian noise, we do see the original model having a higher entropy difference around layers $2$ through $5$, but this effect is mitigated in later layers. This suggests that fine-tuning on high frequency noise may not change the attention distribution as much.

\subsection{SAE Representation Analysis}

We quantify the similarity between a ViT representations of unseen ImageNet images and their corrupted counterpart by calculating the cosine between the extracted SAE activations when the original and corrupted images are fed into the model. 

For both BatchTopK and Vanilla SAE, we use an SAE expansion factor of 32 and $\lambda $. The BatchTopK SAE additionally uses $\alpha=\frac{1}{32}$ and k=32. Both types of SAEs are trained for 15 epochs with early stopping and an initial learning rate from 1e-3 paired with a Cosine Annealing scheduler. We train the SAEs on a mix of uncorrupted and corrupted images to learn a comprehensive feature dictionary.

\begin{figure}[t]
    \centering
    \includegraphics[width=0.9\linewidth]{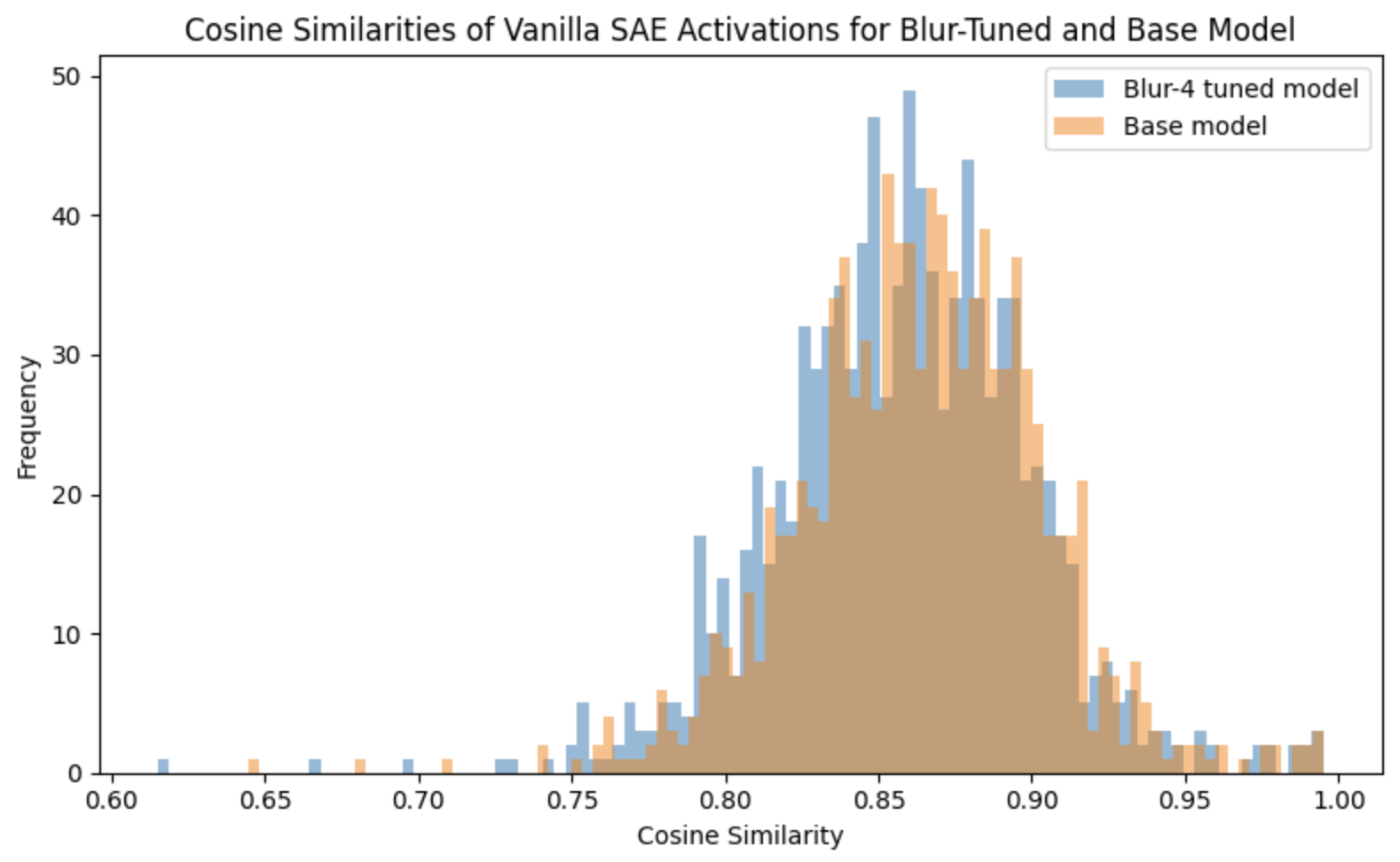}
    \caption{Distribution of cosine similarities of vanilla SAE activations between pairs of original and corrupted images for a blur-4-tuned ViT and the base ViT.}
    \label{fig:vanilla_cos_sim}
\end{figure}

\begin{figure}[t]
    \centering
    \includegraphics[width=0.75\linewidth]{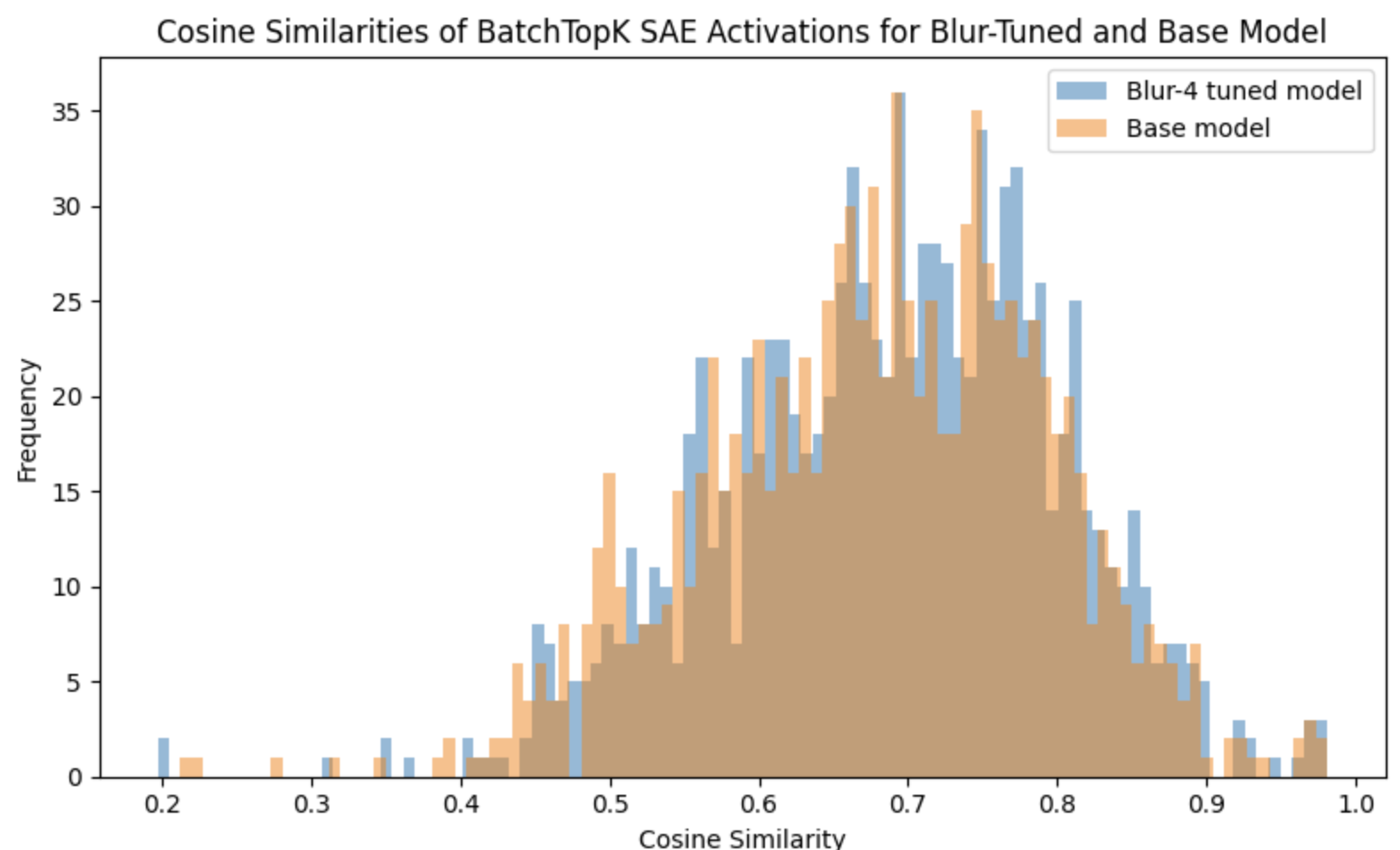}
    \caption{Distribution of cosine similarities of BatchTopK SAE activations between pairs of original and corrupted images for a blur-4-tuned ViT and the base ViT.}
    \label{fig:batchtopk_cos_sim}
\end{figure}

In general, we observe that the distributions are similar between the models further fine-tuned on corrupted data and the base model. \cref{fig:vanilla_cos_sim} and \cref{fig:batchtopk_cos_sim} show the distribution of the cosine similarities between SAE activations for corresponding patches in images and their blurred counterpart in a blur-4-tuned ViT compared to the base ViT.

The nearly-identical distributions of cosine similarities between the two models suggest that the improvement in model robustness gained from adversarial training may not be closely tied to fundamental differences in learned representations of the model.
\section{Discussion and Conclusions}

In this study, we analyze how ViTs process perturbed images. We fine-tune ViT classifiers on common perturbations and compare how the internal processing of these models differs to understand the impacts of adversarial fine-tuning by looking at the class prediction progression across layers, the model's latent representation space, and its attention mechanism. 

We find that adversarial training improves performance and certainty on the adversarial dataset. However, we find that improved performance does not transfer to other families of perturbations, underscoring the importance of comprehensive adversarial training data. Mechanistically, we observe that models fine-tuned on perturbations are more likely to predict the correct answer at an earlier layer than the original model and have more certainty in their answer. 

Future work could investigate how training transfers between different filters of the same family. Future work for the mechanistic analysis could include trying to isolate and control the key differences between the base and the fine-tuned model. 

{
    \small
    \bibliographystyle{ieeenat_fullname}
    \bibliography{main}
}

\end{document}